\documentclass[conference]{IEEEtran}
\usepackage{times}

\usepackage[numbers]{natbib}
\usepackage{multicol}
\usepackage[bookmarks=true]{hyperref}
\usepackage{graphicx}
\usepackage{capt-of}
\let\labelindent\relax
\usepackage{enumitem}
\usepackage[table,xcdraw]{xcolor}
\usepackage{multicol}
\usepackage{multirow}
\usepackage{CJKutf8}
\usepackage{booktabs}
\usepackage{comment}
\usepackage{balance}
\usepackage{lscape}
\newcommand{\pquotes}[1]{\textcolor[gray]{0.25}{\textit{#1}}}

\def\eg{\emph{e.g., }} 
\def\ie{\emph{i.e., }} 

\usepackage{etoolbox}
\newcommand{\insertfig}{\vspace{-1em}\includegraphics[width=\linewidth]{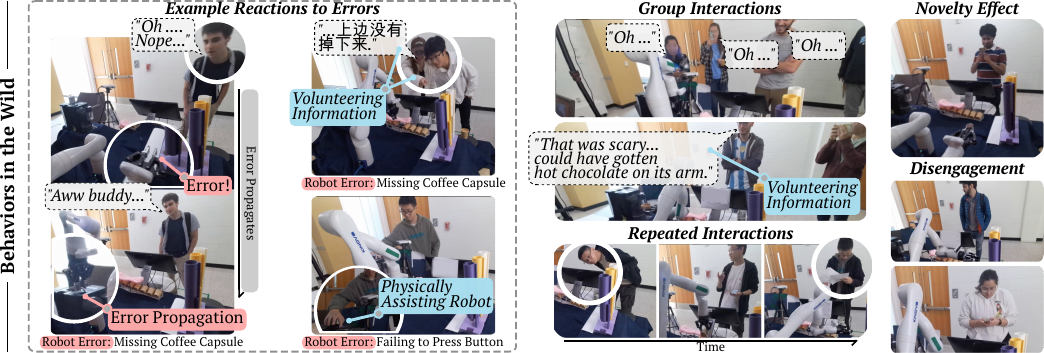}\captionof{figure}{We explore user social signals during in-the-wild interactions with an autonomous coffee-making robot. We show through a field deployment that people reliably express rich but ``noisy''  signals in response to robot errors. We highlight opportunities and challenges for using these cues in real-world HRI.}\label{fig:teaser}}

\makeatletter
\apptocmd{\@maketitle}{\centering\setcounter{figure}{0}\insertfig}{}{}
\makeatother

\begin{document}

\title{Signal or `Noise': Human Reactions to \\Robot Errors in the Wild}

\author{\authorblockN{Maia Stiber}
\authorblockA{Microsoft Research\\
Redmond, WA, USA\\
maiastiber@microsoft.com}
\and
\authorblockN{Sameer Khan}
\authorblockA{Johns Hopkins University\\
Baltimore, MD, USA\\
sameerKhan@outlook.com}
\and
\authorblockN{Russell Taylor}
\authorblockA{Johns Hopkins University\\
Baltimore, MD, USA\\
rht@jhu.edu}
\and
\authorblockN{Chien-Ming Huang}
\authorblockA{Johns Hopkins University\\
Baltimore, MD, USA\\
chienming.huang@jhu.edu}}

\maketitle

\begin{abstract}
In the real world, robots frequently make errors, yet little is known about people's social responses to errors outside of lab settings. Prior work has shown that social signals are reliable and useful for error management in constrained interactions, but it is unclear if this holds in the real world---especially with a non-social robot in repeated and group interactions with successive or propagated errors. To explore this, we built a coffee robot and conducted a public field deployment ($N = 49$). 
We found that participants consistently expressed varied social signals in response to errors and other stimuli, particularly during group interactions. Our findings suggest that social signals in the wild are rich (with participants volunteering information about the interaction), but ``noisy.'' We discuss lessons, benefits, and challenges for using social signals in real-world HRI.
\end{abstract}

\IEEEpeerreviewmaketitle

\section{Introduction}
In lab settings, social signals---verbal and nonverbal cues people naturally express during interaction---have been proven to be a good and reliable indicator of robot errors~\cite{bremers2023using,kontogiorgos2021systematic,spitale2023longitudinal,stiber2023using}. Models of these signals, when integrated into larger robotic systems, enable flexible and automatic error detection~\cite{stiber2025robot}. However, this has been in the lab with single users and single interactions, though a few studies have begun exploring social robot interaction ruptures~\cite{spitale2023longitudinal}. 

It is not evident how these signals manifest outside the lab with a non-social robot. Prior work has not examined responses to errors during group and repeated interactions in public spaces. Our questions are: (1) \emph{how do people react to robot errors in the wild?}; \emph{(2) how do people react during interactions in the wild with a non-social robot across successive errors, repeated interactions, and in group interactions?}; and \emph{(3) what opportunities and challenges arise for using social signals to manage errors in real-world systems?}

To answer these questions, we conducted a three-week public field deployment. First, we developed an autonomous coffee robot field deployment and data collection system to (1) collect data, (2) autonomously provide a service that is commonly experienced in the real world, and (3) safely operate in the face of user interference and without experimenter supervision. Then, we ran the deployment and collected 39 interactions with 49 unique participants.

We found that individuals and groups responded socially to robot errors in the wild across repeated interactions (Fig.~\ref{fig:teaser}), consistently displaying diverse verbal and non-verbal social signals. These were triggered not only by the robot and its errors (or absence of) but also by people, phones, and environmental factors. Group interactions produced more complex social signals, with a higher density of relevant and irrelevant (``noisy'') signals. Thus, social signals serve as a modality for participants to volunteer information about the interaction and  reflect participants' shifting mental models across multiple errors and interactions. This work's main contributions are:
\begin{itemize} [leftmargin=*]
    \item \textbf{Autonomous Data Collection and Robotic Deployment Platform} contextualized as a coffee making robot\footnote{See Appendix 1 for link to system's repository.}.
    \item \textbf{Empirical understanding} of individual and group social responses to robot errors in the wild (\ie during unconstrained interactions, repeated interactions, successive errors).
    \item \textbf{Implications of} and \textbf{Lessons about} using social signals in the wild and field deployments with robots.
\end{itemize}

\section{Background}

\subsection{HRI in the Wild}
When robots are deployed in public, they must interact with diverse groups in dynamic, unpredictable contexts with few limitations on interactions. HRI studies are typically held in lab settings to allow for controlled experimentation; however, research in HRI in the wild reveals phenomena that controlled environments would not be able to yield~\cite{innes2021experimental}, such as different user attention and gaze shift patterns~\cite{chan2025field,lee2024consumer}. While lab studies are ``cleaner,'' they do not account for the unpredictable world where robots will be used~\cite{jung2018robots,sabanovic2006robots}. 

Example field studies have used a robot receptionist~\cite{ben2019early,michalowski2006spatial}, conversational robot in a mall~\cite{koike2025what}, and commercially deployed coffee robots~\cite{lee2024consumer} to explore user engagement. They have explored behaviors around trashcan robots~\cite{brown2024trash,bu2025making}, commercially deployed robots~\cite{chan2025field}, robotic well-being coaches~\cite{spitale2023robotic}, 
social robots in public spaces~\cite{schiffmann2025predicting}, and people's willingness to assist social robots in a mall~\cite{yamada2024qualitative} or with coffee making when asked by a robot~\cite{huttenrauch2003help}. Other systems include shopkeeper~\cite{edirisinghe2024field} and security guard robots~\cite{edirisinghe2024security}.

Within the domain of robot errors in the wild, research has shown that robots make mistakes that affect user interactions~\cite{andrist2017what}. Field studies with social robots have developed taxonomies of interaction patterns affected by errors~\cite{koike2025what}, and explored social signals triggered by interaction ruptures over four weeks~\cite{spitale2023longitudinal}. Field observations of coffee robots in Japan report that workers frequently encounter errors, often daily, and need to figure out how to fix them on their own~\cite{kamino2023coffee}, which can be challenging for end-users.

\subsection{Use of Social Signals in HRI}
User social signals, both explicit and implicit, convey information about users, robot actions, tasks, users' mental models of the robot, and social dynamics within interactions~\cite{bremers2023using,duric2002integrating,tabrez2020survey,vinciarelli2008social} and facilitate communication~\cite{sauppe2014social}. These occur not only in social interactions but also in physical-based interactions~\cite{gucsi2025hrisense,mirnig2017err} and in the wild in social interactions~\cite{koike2025what}. While these signals can be ambiguous~\cite{hassin2013inherently,parreira2024bad,stiber2022modeling}, they also provide information about the interaction. They have been used for detecting confusion~\cite{li2025hri,scherf2024you},  engagement~\cite{ben2019early},  preference~\cite{candon2023nonverbal}, intention~\cite{belcamino2024gaze}, and robot errors~\cite{bremers2023using,kontogiorgos2021systematic,mirnig2017err,spitale2023longitudinal,stiber2025robot}. 

\emph{\textbf{Social Signal Responses to Robot Errors}.}
Social signals have  been explored for robot error management. Robot errors are defined as a misalignment between the user's mental model of the interaction and the robot's actions; that misalignment elicits social signals~\cite{cao2025err,stiber2023using} that have been shown to be exhibited reliably and more frequently when errors are present~\cite{cahya2019static, giuliani2015systematic,mirnig2015impact,trung2017head}. Users respond to errors across multiple modalities, such as  gaze~\cite{aronson2018gaze,severitt2024communication,wolf2021gaze,tabatabaei2025gazing}, facial expressions~\cite{bremers2023bystander,parreira2024bad,stiber2022modeling}, 
verbalizations~\cite{kontogiorgos2021systematic}, and body movements~\cite{giuliani2015systematic}. These reactions shift over time as the user interacts with the robot~\cite{spitale2023longitudinal}; repeated errors affect signals expressed~\cite{liu2025done}. By using social signals as indicators of errors, we are treating ``humans as sensors'' to evaluate robots~\cite{lewis2009using}.

In pursuit of this goal, datasets capturing these signals with both social and non-social robots have been collected~\cite{bremers2023bystander,stiber2023using} and modeling grand challenges for these signals have been created~\cite{cao2025err,spitale2024err}. Modeling shows these signals' utility for error detection~\cite{kontogiorgos2021systematic,ravishankar2024zero,severitt2024communication,stiber2022modeling,stiber2023using,tabatabaei2025real}; their integration into robotic systems allows for flexible, proactive error detection, which users perceive more favorably than reactive approaches~\cite{stiber2025robot}.

However, these works have been in controlled lab settings or with social robots in the field. To the best of our knowledge, little work has been done exploring social signal responses to robot errors in the wild for non-social robots during group interactions, repeated interactions, and successive errors.

\section{Field Deployment and Data Collection System}
\label{sec:system}

\begin{figure*}
    \centering
    \includegraphics{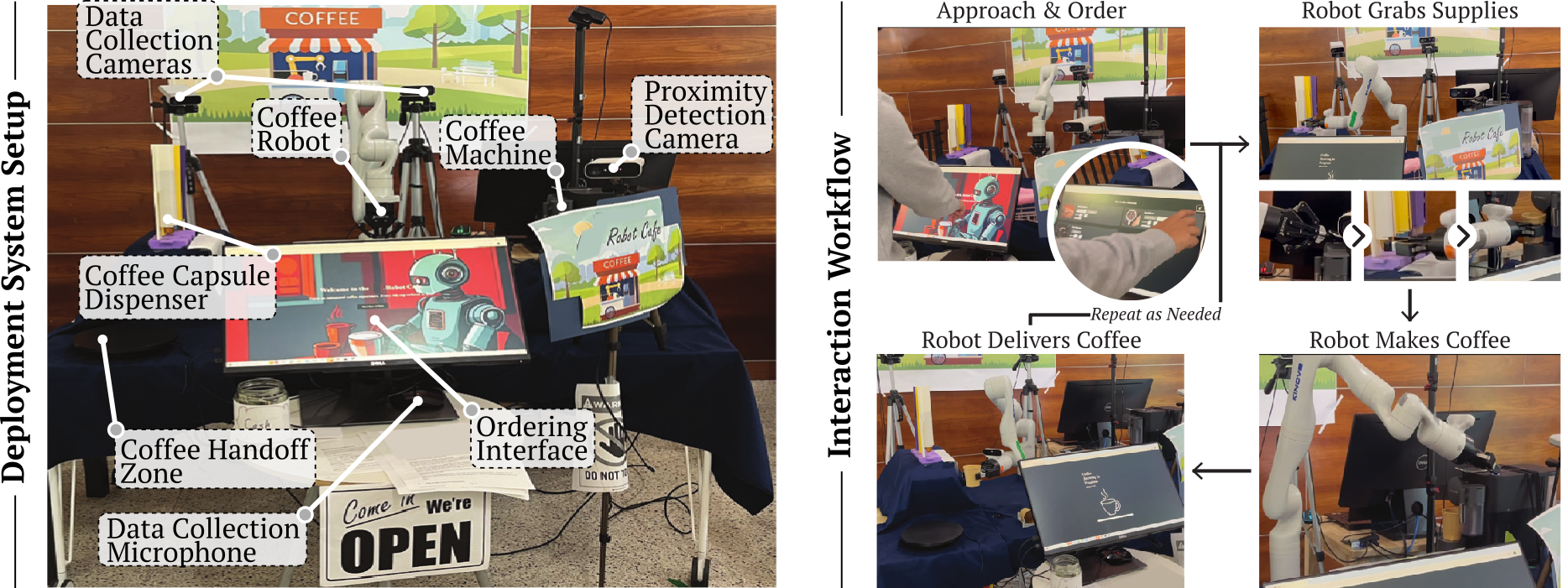}
    \caption{Coffee robot system physical setup (left) and interaction workflow (right).}
    \label{fig:coffee-robot-workflow}
\end{figure*}

\begin{figure}
    \centering
    \includegraphics[width=\columnwidth]{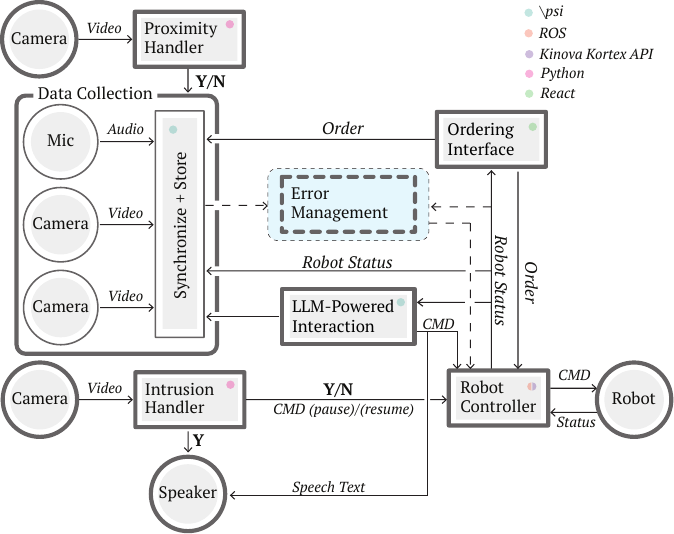}
    \caption[Coffee robot system diagram.]{Coffee robot system diagram. Dashed lines and boxes show how an error management system using social signals could be integrated into it.}
    \label{fig:coffee-robot-system}
\end{figure}

To collect reactions in the wild to a robot that offers an ``everyday'' physical service, we developed a field deployment robotic system contextualized as a coffee making robot.
The design goals for this system were to enable autonomous service capabilities in a public space, handle hours of operation without intervention, respond to user input, make multiple cups of coffee safely, and collect data after user consent.

\subsection{System Overview}
The system (physical layout and workflow in Fig.~\ref{fig:coffee-robot-workflow}; system diagram in Fig.~\ref{fig:coffee-robot-system}) enables users to order coffee for the robot to make. Ordering is via touch screen and video and audio data are automatically collected. This system is autonomous and capable of long-term deployment without experimenter intervention.
The coffee-making process (Fig.~\ref{fig:coffee-robot-workflow}-right) involves the robot turning a coffee machine on, opening the lid, placing the requested capsule in the machine, placing an empty cup in the machine, pressing a button to brew, and delivering the cup of coffee in the designated delivery zone.

\emph{\textbf{Physical Layout and Hardware.}}
The system uses a Kinova Gen3 controlled via ROS with a Kinova API; the user interface is built in React. The system receives inputs via a microphone, three Kinect cameras, and one webcam. The microphone and two Kinect cameras are used for data collection. The third Kinect camera detects when someone is in the interaction area and the webcam is used to determine whether a user has encroached into the robot's operating space. A capsule coffee machine is the coffee making mechanism, while custom-designed capsule and cup dispensers allow the robot to grasp necessary items. A turntable is used as a delivery mechanism to users, supporting multiple coffee orders. 

The physical layout (Fig.~\ref{fig:coffee-robot-workflow}-left) has data collection cameras placed for a wide field of view to record group interactions. To enable natural interactions and not prime users that the robot is dangerous or error-prone, the system has an open layout with no barrier between the user and the robot. Users can reach into the table space, though signs dissuade tampering. For safety, the system has an \emph{Intrusion Detector} to stop the robot when a user enters its operating space.

\subsection{Software} The system is built on \textbackslash psi~\cite{bohus2021platform} and includes (Fig~\ref{fig:coffee-robot-system}): Proximity Handler, Ordering Interface, Data Collection, Intrusion Detector, Robot Controller, and LLM-Powered Interaction\footnote{Not used in deployment. See appendix for description of this component.}. Inter-component communication is via \textbackslash psi and sockets.

\subsubsection{Proximity Handler} \label{sec:proximity-handler}
The \emph{Proximity Handler} takes input from a Kinect camera to determine if someone entered the robot's ``interaction area:'' a pre-defined taped area around the system to denote the recording area. This is done by using the camera's depth image to detect if a new object has entered the field of view, difference between the current depth image and an ``initial'' one (taken during system set up). If a new object is detected within a threshold distance, the system assumes that it is a person.
Video from this camera is not stored, so that people will not be recorded without their consent. 

When a user is detected, this component signals \emph{Data Collection} to start recording (turning the system from idle to running). If the system was ``running'' and the component determined the person left, a countdown starts and when done, this component tells \emph{Data Collection} to stop recording. If the robot is still making coffee (\ie the user left after ordering), the robot finishes the order then initiates the countdown.

\subsubsection{Data Collection}
The \emph{Data Collection} Component records and stores data (audio and video) when a user has approached and consented. It also stores order information from the \emph{Ordering Interface} and robot status information from the \emph{Robot Controller}.
The audio stream and robotic status can also be fed from this component to \emph{LLM-Powered Interaction} when that component is toggled on and requested.
For future extensibility (\eg using data in real-time for an error management component), this component can provide collected information to other components---as shown in Fig.~\ref{fig:coffee-robot-system}. 

\subsubsection{Ordering Interface}
Through the \emph{Ordering Interface}, users can provide consent, select coffee flavor, size and quantity, and pay. Once the place order button is tapped, this component sends the order to \emph{Data Collection} and the \emph{Robot Controller}. Orders are also stored on the backend to track supply levels. If the system is running low, the user interface displays a service alert that disappears once the experimenter resupplies. When this component receives a ``done order'' notification from the \emph{Robot Controller}, it sends follow-up survey email to the user and resets for the next interaction. 

\subsubsection{Intrusion Detector}
The \emph{Intrusion Detector} determines if someone is reaching into the robot's operating region (pre-defined) to prevent collisions. This is necessary since the system is not placed behind a divider.
It detects this using MediaPipe hands landmark detector and determines if a hand intersects with the region. This assumes that any human intrusion into the robot's space would involve at least a hand. If a hand is detected, then the \emph{Intrusion Detector} sends a pause command to the \emph{Robot Controller} and the speaker verbally warns the user to move their hand. If the hand is removed, then it sends a resume command to the \emph{Robot Controller}.

\subsubsection{Robot Controller}
The \emph{Robot Controller} communicates with the robot and with the other components about the robot. It receives orders from the \emph{Ordering Interface} and executes corresponding actions. It also receives pause and resume commands from the \emph{Intrusion Detector}.
The robot sends notifications to the robot controller when it is moving and when the order is done. This robot status information is communicated to the rest of the system. Additionally, the controller can introduce pre-defined errors into the coffee making execution at a preset rate.

\section{Field Deployment in a Public Space}
We conducted a field deployment with our system to explore interactions in the wild (unconstrained interactions in an unconstrained environment) with a potentially faulty robot. Our goals were three-fold:
(1) understand how people react to robot errors in a public space in the wild;
(2) explore what social signals are expressed during group interactions, repeated interactions, and across successive and propagated errors; and 
(3) determine opportunities and challenges of using social signals as input for error management in deployed robotic systems. Prior work shows the promise of using social signals for error detection (\eg \cite{kontogiorgos2021systematic,stiber2025robot}), but these studies, to our knowledge, are largely contextualized in lab settings.

\subsection{Study Design}
Our study allowed unconstrained interactions in unconstrained environments; people could freely approach and interact. Experimenters were not present during deployment, except for checking in to resupply when participants were not present. The system was deployed every day for about three weeks, six hours at a time. Users had the option of three flavors (medium roast, dark roast, hot chocolate) and two sizes (espresso, lungo) and could order up to five cups at once. 

\subsubsection{Deployment Setup}
The system was deployed in a university building hallway near a seating area, allowing those sitting and walking by to interact with it at their discretion (Fig.~\ref{fig:coffee-robot-workflow}). We taped an area around the system, defined by settings in \emph{Proximity Handler} (Section~\ref{sec:proximity-handler}), that visually indicated to people that they would be recorded if they stepped into that region (recordings were kept if participants filled out a written consent form). In addition, we placed signs next to the interaction region and on doors leading to the hallway to provide information about the taped area and the data that was being recorded. The \emph{LLM-Powered Interaction} component \emph{was not} toggled on so the robot would be non-social.

\subsubsection{Error Manipulation}
There were four possible pre-programmed errors: (1) failing to grab the cup after coffee had been made, (2) not putting a coffee capsule in the machine, (3) not putting a cup under the coffee machine, and (4) not pressing the coffee making button. These pre-determined errors were set to randomly trigger with a maximum rate of one per coffee. Not all interactions with the robot resulted in a triggered error as we wanted to encourage repeat interactions. Additionally, the robot would automatically resolve all triggered errors so that users could get the drink they paid for. Inevitably, with any deployed robot, there were unexpected (by the developer) errors, so there could be more than one error per order. Responses to those were also captured. Those errors, however, were not automatically resolved.

\subsection{Study Procedure}
If a person wanted participate, they would step over a taped line in front of the robot; a sign indicated that by doing so, they were willing to be recorded. They then interacted with the touch screen ordering interface where they first provided explicit consent to the study. After consenting, they would create, pay for and place their order.
Participant payment was not tracked (\ie based on the honor system). The robot then went about making the order while the participant watched. After the robot was done, the participant could retrieve their order and a post-study survey was automatically sent to the email address they provided to learn more about their experience. This study was approved by the Institutional Review Board.

\subsection{Dataset Statistics and Participants}
We collected a total of 39 \emph{complete} interactions (\ie participants initiated interaction and placed an order) comprised of 13 group interactions---two to seven people in size---and 26 single person interactions. Fourteen of those interactions were repeat interactions. Repeat participants were identified manually via video and provided email addresses. Two participants interacted individually and later in a group; there were no groups repeated. In actuality, there were over 47 initiated interactions with the robot (\ie combination of \emph{complete} and \emph{incomplete} interactions, where participants approached but did not place an order). The \emph{incomplete} interactions were not considered part of the dataset and not included in the following data analysis. 

Nine interactions involved non-English speech. Of these, seven were fully or partially in Chinese (we were able to transcribe and annotate), one was English and another language (only the English portion was annotated), and one was in another language (audio excluded due to lack of transcription). Occasionally, participants interacted with passersby, only participants' data was analyzed and passerby audio was excluded.
We collected and analyzed 211 min of data with an average interaction session length of 5.43 min ($SD = 1.70$). 

There were 49 unique participants in the dataset of which seven participants initiated repeat interactions, ranging from one to seven repetitions. Participants ordered from one to three coffees per interaction. Due to the nature of the field study, we could not collect users’ biographical information.

\subsection{Dataset Analysis and Metrics}
We conducted thematic analysis on user behavior. Two coders annotated interaction video and audio using both inductive and deductive coding and agreed upon annotations. Codes were derived from our research goals and new codes were added as patterns in behavior were observed. The coding process involved analyzing user behavior during the general interaction and errors separately (following similar methodology to \cite{koike2025what} for interaction phases). Once coding was completed, the researchers examined patterns to develop metrics along with the themes. For repeated interaction analysis, groups with repeat individuals were counted as repeated, but analysis focused on the repeated individual.

\subsubsection{Codes} There were three high-level categories of codes: Robot Error Codes (predefined based on research goal), Behavioral Codes (established through open coding and predefined codes based on research goal), and Verbal Level of Relevancy and Information Codes (open coding). These codes were used to describe both user and robot behavior during an interaction. See Appendix 2 for codebook.
The high-level code categories were:
\begin{itemize}[leftmargin=*]
    \item \textbf{Robot Error Codes} were used to indicate whether an interaction contained an error and, if so, used to identify the start (first time when from video it could be recognized).
    \item \textbf {Behavioral Codes} were used to annotate the presence and stimuli of different user behaviors (verbal and non-verbal) and reactions during error instances and general interaction---outside of error execution---as well as start and end of user reactions to errors and speech valence. User behavior presence codes were adapted from \cite{koike2025what,stiber2020not} and additional open coding. High-level categories of behavioral signals were facial, verbal, and body movement.
    \item \textbf{Verbal Level of Relevancy and Information Codes} were used to annotate speech at the sentence level, assessing the relevance and the amount of information provided about the robot and its errors (evaluated separately for robot-related and error-related content). These were numerical codes, with  $>0$ indicated relevant information and $>1$ indicated relevant \emph{and} volunteered information. Video was used alongside the transcripts to contextualize speech, as many sentences relied on deictic expressions and demonstrative pronouns.
\end{itemize}

\subsubsection{Metrics}
We then developed metrics to provide further analysis alongside the qualitative codes. These metrics were:
\begin{itemize}[leftmargin=*]
    \item \textbf{User Reaction Percentage}: Derived from Robot Error Codes, the percent of errors reacted to by participants. 
    \item \textbf{User Reaction Time (s)}: The reaction time to the error's manifestation; the time between when the coder saw the first possible moment that a participant could have recognized the error and the initial social signal seen when participants reacted to the error. Derived from the Behavioral Codes.
    \item \textbf{User Reaction Duration (s)}: The length of initial visible reaction to an error, calculated as \textit{reaction end} (when participant's behavior returns to their norm in the video) minus \textit{reaction start}. Derived from Behavioral Codes.
    \item \textbf{Percentage of Robot Errors with Volunteered Information}: Derived from Verbal Level of Relevancy and Information Codes, the percent of robot errors that participants provided additional information via speech. Sentences rated as $\geq 2$ indicated that the user provided more information about the error other than just occurrence. Each such sentence was mapped to an error to calculate percentage.
    \item \textbf{Rate of Irrelevant Speech (sentences/min)}: This quantified the average amount of speech during an interaction that was not relevant to the robot or task. Derived from Verbal Level of Relevancy and Information Codes, it was the number of sentences spoken rated as 0 for the robot-related codes divided by the length of interaction in minutes.
    \item \textbf{Rate of Volunteered Speech Information (sentences/min)}: Derived from Verbal Level of Relevancy and Information Codes, quantified the average number of sentences (rated $\geq2$) per minute that provided additional, volunteered information about the robot, task, or user perception.
\end{itemize}

\section{Findings}

We analyzed social signals and identified behavioral patterns from in the wild interactions during robot errors, general interaction with the robot, and across interactions. ``I$X$'', where $X$ is a number, refers to a particular interaction in the dataset.

\emph{\textbf{Robot Error Statistics:}}
Of the 39 interactions, 25 interactions (single: 15; group: 10) had errors and 14 interactions were error-free---totaling 43 errors experienced by participants during deployment. Thirteen interactions had successive errors (more than one). There were nine types of errors (not all pre-planned) including: the robot failing to grab a coffee capsule, improperly putting the capsule in the coffee machine, not pressing the make coffee buttons, and being nonresponsive.

There were three robot behaviors that were not errors but were considered errors by some participants: grabbing a cup tightly (by design, so the robot could reliably grab the cups), grabbing two cups (by design, to prevent filled cups from being too hot), and pulling on the capsule dispenser hard. While most errors were triggered by the robot, some were caused by incorrect human action when participants tried to help the robot. Because the robot could not recognize this assistance, and participants did not realize their actions were incorrect, these mistakes were attributed to the robot.

\subsection{User Reactions to Robot Errors in the Wild}

\begin{table}
\centering
\renewcommand{\arraystretch}{1.05}
\caption{User Social Signals Percentage, verbal and non-verbal.}
\label{tab:ss-frequencies}
\resizebox{\columnwidth}{!}{%
\arrayrulecolor{black}
\setlength{\arrayrulewidth}{0.7pt}
\begin{tabular}{ccllll}
\textbf{} &
  \multicolumn{1}{l}{} &
   &
   &
  \begin{tabular}[c]{@{}l@{}}Reaction to Error\\ (\% of errors)\\ \emph{Total Count: 43}\end{tabular} &
  \begin{tabular}[c]{@{}l@{}}Reaction during\\ General Interaction\\ (\% of interactions)\\ \emph{Total Count: 39}\end{tabular} \\ \cmidrule{3-6} 
 &
   &
  \multicolumn{1}{c}{\cellcolor[HTML]{EFEFEF}} &
  \multicolumn{1}{l|}{\cellcolor[HTML]{EFEFEF}\textit{Robot}} &
  \multicolumn{1}{c}{\cellcolor[HTML]{EFEFEF}14\%} &
  \multicolumn{1}{c}{\cellcolor[HTML]{EFEFEF}10\%}\\
 &
   &
  \multicolumn{1}{c}{\cellcolor[HTML]{EFEFEF}} &
  \multicolumn{1}{l|}{\cellcolor[HTML]{EFEFEF}\textit{Self}} &
  \multicolumn{1}{c}{\cellcolor[HTML]{EFEFEF}7\%} &
  \multicolumn{1}{c}{\cellcolor[HTML]{EFEFEF}8\%} \\
 &
   &
  \multicolumn{1}{c}{\cellcolor[HTML]{EFEFEF}} &
  \multicolumn{1}{l|}{\cellcolor[HTML]{EFEFEF}\textit{Group}} &
  \multicolumn{1}{c}{\cellcolor[HTML]{EFEFEF}51\%} &
  \multicolumn{1}{c}{\cellcolor[HTML]{EFEFEF}33\%} \\
 &
  &
  \multicolumn{1}{l}{\cellcolor[HTML]{EFEFEF}} &
  \multicolumn{1}{l|}{\cellcolor[HTML]{EFEFEF}\textit{Passerby}} &
  \multicolumn{1}{c}{\cellcolor[HTML]{EFEFEF}7\%} &
  \multicolumn{1}{c}{\cellcolor[HTML]{EFEFEF}15\%} \\
  &
  &
  \multicolumn{1}{l}{\multirow{-5}{*}{\cellcolor[HTML]{EFEFEF}Talking to ...}} &
  \multicolumn{1}{l|}{\cellcolor[HTML]{EFEFEF}Total} &
  \multicolumn{1}{c}{\cellcolor[HTML]{EFEFEF}49\%} &
  \multicolumn{1}{c}{\cellcolor[HTML]{EFEFEF}54\%} \\
 &
   &
  \multicolumn{1}{c}{} &
  \multicolumn{1}{l|}{\emph{Laughing}} &
  \multicolumn{1}{c}{35\%} &
   \multicolumn{1}{c}{33\%}\\
 &
  \multirow{-6}{*}{\rotatebox[origin=c]{90}{Verbal}} &
  \multicolumn{1}{l}{\multirow{-2}{*}{Non-Lexical Utterances}} &
  \multicolumn{1}{l|}{\emph{Misc.}} &
   \multicolumn{1}{c}{42\%}&
   \multicolumn{1}{c}{28\%}\\ \cline{3-6}
 &
 &
      \cellcolor[HTML]{EFEFEF}Smiling &
  \multicolumn{1}{l|}{\cellcolor[HTML]{EFEFEF}} & \multicolumn{1}{c}{\cellcolor[HTML]{EFEFEF}60\%}
   & \multicolumn{1}{c}{\cellcolor[HTML]{EFEFEF}62\%}
   \\
 &
   &
  Shocked Expression &
  \multicolumn{1}{l|}{} &
   \multicolumn{1}{c}{23\%} &
   \multicolumn{1}{c}{3\%}\\
 &
  \multirow{-3}{*}{\rotatebox[origin=c]{90}{Facial}} &
  \cellcolor[HTML]{EFEFEF}Grimacing & 
  \multicolumn{1}{l|}{\cellcolor[HTML]{EFEFEF}} & \multicolumn{1}{c}{\cellcolor[HTML]{EFEFEF}21\%}
   & \multicolumn{1}{c}{\cellcolor[HTML]{EFEFEF}0\%}
   \\ \cline{3-6}
 &
   &
  \multicolumn{1}{l}{Gesturing with Hands} &
   \multicolumn{1}{l|}{}&
   \multicolumn{1}{c}{30\%}&
   \multicolumn{1}{c}{38\%}\\
 &
   &
  \multicolumn{1}{l}{\cellcolor[HTML]{EFEFEF}Leaning in} &
   \multicolumn{1}{l|}{\cellcolor[HTML]{EFEFEF}} &
  \multicolumn{1}{c}{\cellcolor[HTML]{EFEFEF}47\%} &
  \multicolumn{1}{c}{\cellcolor[HTML]{EFEFEF}41\%} \\
  &
   &
   Looking away &
   \multicolumn{1}{l|}{}&
   \multicolumn{1}{c}{23\%} &
  \multicolumn{1}{c}{56\%} \\
 &
  &
   \cellcolor[HTML]{EFEFEF}&
  \multicolumn{1}{l|}{\cellcolor[HTML]{EFEFEF}\emph{Around}} &
   \multicolumn{1}{c}{\cellcolor[HTML]{EFEFEF}14\%} &
  \multicolumn{1}{c}{\cellcolor[HTML]{EFEFEF}10\%}\\
 &
   &
  \multirow{-2}{*}{\cellcolor[HTML]{EFEFEF}Walking...} &
  \multicolumn{1}{l|}{\cellcolor[HTML]{EFEFEF}\emph{Away}} &
   \multicolumn{1}{c}{\cellcolor[HTML]{EFEFEF}7\%}&
   \multicolumn{1}{c}{\cellcolor[HTML]{EFEFEF}10\%}\\
 &
   &
  Shaking Head &
  \multicolumn{1}{l|}{} &
  \multicolumn{1}{c}{9\%}&
   \multicolumn{1}{c}{3\%}\\
 &
   &
  \multicolumn{1}{l}{\cellcolor[HTML]{EFEFEF}Physically Assisting Robot} &
  \multicolumn{1}{l|}{\cellcolor[HTML]{EFEFEF}} &
   \multicolumn{1}{c}{\cellcolor[HTML]{EFEFEF}44\%}&
   \multicolumn{1}{c}{\cellcolor[HTML]{EFEFEF}0\%}\\
 \multirow{-16}{*}{\rotatebox[origin=c]{90}{\textbf{Social Signal Modalities}}}&
  \multirow{-8}{*}{\rotatebox[origin=c]{90}{Body Movement}} &
  Freezing &
  \multicolumn{1}{l|}{} &
   \multicolumn{1}{c}{14\%} &
   \multicolumn{1}{c}{0\%}\\ \midrule \midrule
 & 
   &
    Recording Robot &
  \multicolumn{1}{l|}{} &
   \multicolumn{1}{c}{7\%}&
   \multicolumn{1}{c}{44\%}\\
 &
  \multirow{-2}{*}{\rotatebox[origin=c]{90}{Phone}} &
  Using Phone &
  \multicolumn{1}{l|}{} &
   \multicolumn{1}{c}{5\%}
   & \multicolumn{1}{c}{38\%}
  
\end{tabular}%
}
\end{table}

Social signals were reliably exhibited in response to errors in the wild, though they manifested in a wide variety of ways. Their expression reliability matched in lab results~\cite{cuadra2021my,mirnig2017err,stiber2023using}; however, the modality range was broader, as interactions were unconstrained (\eg \emph{walking away, talking with passersby}).

\subsubsection{Error Reaction Statistics}
Of the errors collected in this dataset, 97.67\% elicited user reactions. The most common were smiling, talking, leaning in to get a better view, and physically assisting the robot. Table~\ref{tab:ss-frequencies} presents the percent of errors that triggered each behavioral response. The average \emph{user reaction time} was $18.81s$ ($SD = 36.56$) and average initial \emph{user reaction duration} was $9.48s$ ($SD = 6.04$). Reactions typically began with implicit signals, followed by explicit ones, consistent with prior findings~\cite{stiber2020not}. Participant reactions to errors extended beyond the initial reaction as participants discussed errors and reacted to error propagations. 

When it came to verbal responses to errors, the dataset contained 135 sentences spoken that were robot error related of which 66.40\% provided additional information about the errors---more than presence. As a result, the \emph{percentage of robot errors with volunteered information} was 58.14\%. Examples of additional information are \pquotes{``More importantly, it just bonked a camera.''} (I7) and \pquotes{``hmm hot water''} spoken when the participant received their coffee order (I36).

\subsubsection{Reactions to \textbf{Error Propagation}}
Each further robot post-error action incurred additional participant reaction. There were 22 errors not immediately recovered from---either due to system limitations or because they were unplanned. These unmanaged errors had the potential to propagate in the task. For example, from I26, the robot dropped a coffee capsule on the floor and the propagation chain following the error was robot miming placing the capsule in the coffee machine, closing the machine clip, and brewing but making water. Of the propagated errors, 63.64\% elicited further reactions, per error one to three additional reactions for a total of 19 additional reactions. These reactions became more explicit as the error propagated, more direct and incorporated more information---involving active debugging (\eg \pquotes{``I think it didn't press hard enough so it didn't puncture the pod.''} (I18)). 64.29\% of the propagated errors that elicited reactions had more explicit reactions. For example, an initial reaction of \pquotes{``Oh no''} became \pquotes{``He dropped the K cup... Well, I don't think it's going to brew if there's no K cup in there, right?''} (I26). Another example had the participant \pquotes{leaning in} when the robot failed grab a pod, later becoming \pquotes{leaning in, shocked expression, and gesturing with hands} (I6) when the robot delivered only water.

\subsubsection{Reactions to \textbf{Successive Errors}}\label{sec:successive}
Additionally, 13 interactions contained more than one error (up to four). Successive errors caused shifts in both user engagement and reaction valence. Participants' responses to multiple errors varied from disengaging (46.15\% of interactions)---looked at phone, read consent forms, or walked away---to physically stepping in to help the robot (69.23\% of interactions). An example of physical assistance was grabbing a cup for the coffee machine (I29). Successive errors also led to more frequent negative valence speech (53.84\% of interactions). Participants speculated about future robot mistakes or jeered at the robot. Examples are \pquotes{``I'm just waiting for it to just pick it up and like drop it.''} (I26) and \pquotes{``\begin{CJK*}{UTF8}{gbsn}它有点傻吧\end{CJK*} [It [the robot] is a little stupid].''} (I29).

\subsection{Reactions across \textbf{Repeated Interactions}} 
Multiple interactions with the coffee robot affected users' responses. Over repeated interactions, participants tended to stop recording the robot (71.43\% of repeated interaction participants) and disengage (42.8\% of repeated interaction participants)---walk away to let the robot do its business, look at phone, and read consent forms. For participants who experienced errors in prior interactions and continued to in subsequent interactions, their behavioral responses reflected patterns described for successive errors (Section~\ref{sec:successive}). On the other hand, participants who experienced errors in prior interactions but \emph{did not} in later ones reacted to the absence of errors (\eg prior error failing to pick up capsule and reacting to the robot picking it up correctly). This occurred in 42.86\% of repeated interaction participants. Examples of these reactions are \pquotes{``oh it worked this time''} (I20) and \pquotes{leaning in, smiling, and gesturing with hands (thumbs up)} (I35). This pattern was also the case within a single interaction with multiple coffee orders where an error occurred in the first order but not in the second order; participants reacted at the same moment where the error had previously occurred.

\subsection{Reactions in \textbf{Group Interactions}}
Finally, we explored the effects that group versus single-person interactions had on social signals. In group interactions, participants had one more category of stimuli that single person interactions did not: each other. Each member at any point could react to stimuli causing a cascade effect. When looking at social signal response to robot errors, in every group interaction, every member expressed social signals; this was not the case for single-person interactions. For speech, group interactions had a larger \emph{percentage of robot errors with volunteered information} (group: 95.65\%; single: 15.00\%).

The result was a high density of social signals expressed per interaction for groups (\eg speech---group: 68.6 sentences/interaction; single: 3.07 sentences/interaction). Of these sentences, group interactions had a higher \emph{rate of irrelevant speech} (group: 4.18; single: 0.10). Conversely, group interactions also had a higher \emph{rate of volunteered speech information} (group: 2.81; single: 0.20) than individual ones.
Topics of volunteered information ($\%$ of group interactions) were:
\begin{itemize} [leftmargin=*]
    \item errors and their causes (69.23\%): \eg \pquotes{*pointing* ``\begin{CJK*}{UTF8}{gbsn}...没转下来\end{CJK*}[it did not come down]''} (I27),
    \item user perception of robot (69.23\%): \eg \pquotes{``I don't trust it because ... I want to put the cup on first.''} when the robot grabbed the coffee capsule first (I7),
    \item prior interaction experience (15.38\%): \eg \pquotes{``Yesterday there was a shift...Yeah now everything works.''} (I34),
    \item robot operation (38.46\%): \eg \pquotes{``It probably detected that there was nothing there.''} (I7), and
    \item speculating about the robot's actions (38.46\%): \eg \pquotes{``It could’ve put hot chocolate on its arm.''} (I21).
\end{itemize}

\subsection{General User Behavior During Deployment}
Participants expressed social signals, as shown in Table~\ref{tab:ss-frequencies}, not only as a result of robot errors but also correct execution. These reactions were not only to the robot's actions but also to group members (if a group interaction), phones, and bystanders passing by.
Table~\ref{tab:ss-frequencies} shows the behavioral signals and percentage of interactions that these signals were expressed outside of errors. \emph{Smiling} and \emph{talking} were triggered by all four stimuli described above (robot, group members, passersby, and phone), looking away was triggered by passersby, and leaning in was triggered by the robot's actions. Also, recording the robot and general phone usage was very common.

Participants spoke a total of 732 sentences;  51.78\% were relevant to the robot or task. Spoken sentences provided additional information (more than what the robot already knew) 35.11\% of the time.
Examples of  volunteered information were \pquotes{``Oh, that's definitely better than any of us can do.''} (I9) and \pquotes{``I don't think it's going to be able to find its K cup''} (I26).

\section{Discussion}
In this work, we explored user behavior during interactions in a public space with a coffee robot, answering the questions of whether and how people react to robot errors in the wild (R1)---across group and repeated interactions (R2)---and the resulting implications for reactions' use in deployed systems (R3). We showed that participants reliably produced a wide range of verbal and nonverbal social signals to both robot errors---in line with prior lab studies (\eg \cite{mirnig2017err,bremers2023using,stiber2023using,parreira2024bad})---and correct actions. These went beyond indicating presence; they also reflected participants' shifting mental models and perceptions of the interaction. Social signal responses were not just triggered by the robot but also  other people, phones, and the environment---indicating the need for contextualization of these signals for effective usage. The opportunities and challenges of leveraging social signal in HRI are especially amplified for group interactions, where signals are denser, more relevant, and also ``noisier''. Below, we describe  implications for understanding and using social signals in the wild, along with lessons learned from our field deployment.

\subsection{Social Signals in the Wild are \textbf{Information Rich}}
 These social signals carry \emph{rich information} about user perception of robot and interaction. We define them as ``information rich'' because they provide insight into how users interpret the interaction beyond the immediate task; more than the robot needs for its immediate problem, offering info about intent, environment, and user mental models. While our initial focus was reactions to robot errors, it became clear that participants actively contribute \emph{additional information} through implicit and explicit signals, as a result of  both positive (\eg successful robot actions after prior errors) and negative stimuli, and to robot and non-robot stimuli. These reactions provide information that can help robots become more human-aware. 

\subsubsection{Social Signals as a Source of Volunteered Information}
Participants’ reactions were not limited to acknowledging the robot’s actions; they often volunteered information.
This came not only from responses to errors but also after successful recoveries (positive actions; \eg ``it got it right this time''). It extended beyond the robot itself, comparing its performance to humans or attributing outcomes to environmental factors. These signals could allow robotic systems to adapt to users.

For robot errors, these signals contain error recovery information (also seen in lab studies with error reporting~\cite{stiber2025robot}).
Implicit signals convey additional information: \emph{leaning in} and gaze could be used to highlight items pertinent to the errors (as with pointing). Another behavior that holds additional information is participants physically assisting the robot, similarly observed in field deployments with social robots~\cite{yamada2024qualitative}.

\subsubsection{Richness of Social Signals in Group Interactions}
Participants exhibited more social signals during group interactions. Members would debug the error aloud or use errors as conversational grounding. The \emph{high rate of volunteered speech information} shows that errors initiate joint sensemaking.
When one participant reacted, others often responded and elaborated, providing additional interpretations of the interaction.
This not only informed error presence and recovery but also other interaction aspects, such as engagement and user perception.

\subsubsection{Social Signals as Information about Misalignments}
Reactions result from misalignments between participants' mental models and robot actions~\cite{stiber2023using}. Errors are one such action that uniquely highlights that gap; social signals elicited hold information about them.
As participants interact, their mental models continually update, but misalignments can continue, eliciting social signals. This is apparent when analyzing participant behavior during repeated interactions, propagated errors, and successive errors. In repeated interactions, participants expressed social signals both when an error occurred and when it did not in subsequent interactions, as their mental models shifted to expect the error and then shifted again when the robot performed correctly. These signals indicated both negative and positive valenced outcomes of these misalignments.
For propagated errors, users reacted to the initial error and to each subsequent robot action that exacerbated that error (\eg Fig.~\ref{fig:teaser}); the gap between mental models and reality continued to diverge. Social signals were more direct, incorporating both diagnostic information and emotional valence.
Finally, successive errors, like propagated, resulted in \emph{social signals as indicators of worsening trust and disengagement} (\eg users ``called the robot names,'' and walked away). Prior lab work similarly showed that successive errors affect social signals~\cite{liu2025done} and can be used for disengagement detection in the field with a social robot~\cite{ben2019early}.

\subsection{Social Signals in the Wild are \textbf{``Noisy''}}
The social signal data had a high degree of ``noise.'' We use “noisy” to describe social signals that are not directly informative to the current problem, either because they are contextually ambiguous or are in response to something other than the problem at hand. This is not “noise” in terms of sensor error, but rather signals whose meaning is difficult to interpret reliably in real-world settings. They were exhibited at any point, during errors and correct executions; this complicates their use.
The number of sources of stimuli also highlights the complexity of interpreting signals in the wild, exacerbated by their ambiguity.
It is difficult to determine which reactions were caused by the robot's behavior or were influenced by external factors, as well as their valence.

One such confounding factor was the frequent novelty effect displayed. Recording the robot (which disappeared after repeated interactions) alone was in 44\% of interactions. Other indicators of novelty were clapping, smiling, and thumbs up. Novelty impacts the social signals expressed, as limited experience with the robot could increase mental model misalignments. The converse was also true; in 38\% of interactions, participants were distracted by their phones---a modern behavior pattern that might render social signals less useful in real-world HRI. Other work, similarly, found that participants were often preoccupied and not always focused on the robot~\cite{chan2025field,lee2024consumer}---unlike in lab studies.

\subsubsection{Exacerbation of ``Noise'' in Group Interactions}
Group interactions not only enrich information; they also enrich ``noise.''
They shape the behaviors that emerge, with participants paying less attention to the robot~\cite{michalowski2006spatial,sebo2020robots}. This is further shown by the higher rate of irrelevant speech in groups than in individual interactions.

\subsubsection{Duality of ``Noise''}
However, the ``noise'' from social signals is not necessarily irrelevant. What counts as ``noise'' is not absolute, but depends on how the problem space is defined and so is contextual. For example, the volunteered information (\eg \pquotes{``Yeah I'm just afraid the cups going to fall and the water is gonna go everywhere.''} (I24) or \pquotes{giving a thumbs up} (I10)), while perhaps considered ``noise'' for error detection, is valuable signal for other purposes such as error recovery, or assessing user trust or preferences. Thus, framing social signals as inherently noisy or ambiguous is oversimplified.

\subsection{Implications for Signal Use in Robotic Systems}
These findings point to both the promise and challenges of using social signals for adaptive robot behavior in the real world. Their richness suggests feasibility of using these cues to make robots more human-aware, while their noisiness underscores the need for context-aware interpretation.

A main difference between lab and in-the-wild settings is level of control. Lab studies constrain participants: they cannot easily leave, look at their phone, or disengage. In everyday environments, however, signals are embedded within competing stimuli---faces can be blocked by phones (often seen in our dataset), attention can shift, and reactions may not be directly triggered by robot behavior. This complexity in determining signal meaning and combating ``noise'' could be addressed by social signal contextualization. Prior studies in lab settings showed contextualization improves performance~\cite{stiber2025robot,stiber2024uh}, making it even more critical in the wild. Examples of such contextualization could include using scene understanding to determine sources of stimuli to weight based on potential ``noisiness'' of the scene as well as the use of other behavioral cues (such as gaze) to determine from where signals are triggered. Or, one could draw on group dynamics, weighting signals that are shared across multiple members more than isolated cues from a single individual.
Finally, social dynamics added another layer of complexity; effective modeling of social signals may require tailoring models for group interactions rather than using one model for all interactions.

\subsection{Field Deployment Lessons and Limitations}
The real-world environment presented several challenges and insights that informed both our hardware setup and interaction design and will inform future deployments to improve efficiency, robustness, and user engagement.

\subsubsection{Physical System Setup} 
Location was a key factor; our initial site, while convenient, attracted a limited user pool. Sites with higher foot traffic and more diverse participants would address this.
Additionally, frequent redeployment of the system highlighted challenges related to hardware mobility and stability. Our initial layout with multiple tables and external cameras proved cumbersome, restricting deployment locations, prolonging setup and tear down, and requiring frequent robot recalibration. A more compact, well-organized, and secured system would address these issues, particularly when deployed away from storage location.

\subsubsection{Interaction Design}
Several user interaction patterns emerged, highlighting areas for improvement in both data collection methods and sustaining system engagement. One challenge was the low response rate to the post-interaction survey, despite follow-up emails. For future studies that require post-interaction data, alternative methods should be used. However, that does affect the interaction paradigm and realism so there needs to be a balance.

We also found reduction in engagement from repeat users as the deployment progressed.
Future work should explore engagement strategies (\eg small talk using the system's \emph{LLM-Powered Interaction} component). This can create rapport, increasing engagement~\cite{pineda2025see} and forming an ingroup. Ingroup bias can mitigate robot errors~\cite{stiber2022Forging}. Maintaining engagement would have users pay closer attention to the robot (difficult with the presence of phones)---enabling more immediate reactions to errors. Additionally, implementing mechanisms that encourage more return visits or ensure the robot delivers a service people would seek out regardless of whether it is performed by a robot could sustain more repeat interactions.

\section{Conclusion}
Interactions in the wild lead people to produce social signals that are rich but ``noisy.'' These signals often volunteer information and are shaped by robot actions, environmental factors, and evolving users’ mental models. Group interactions amplify their richness and ambiguity, while repeated interactions reveal they reflect shifts in users' models. Our findings illustrate both the promise and challenge of using social signals for real-world HRI, emphasizing the need for context-aware methods that can reliably distinguish useful cues from ``noise.''

\section*{Acknowledgment}
The work was supported by the National Science Foundation award \#2143704. We would like to thank Nadia Kim for building the UI for the system and Zitong Wei for designing and building the capsule and cup dispensers.

\section*{CRediT author statement} MS: Conceptualization, Methodology, Software, Validation, Formal Analysis, Investigation, Data Curation, Writing - Original draft, Writing - Review \& editing, Visualization, and Project Administration.
SK: Software, Validation, Formal Analysis, Investigation, Writing - Original draft, and Writing - Review \& editing.
RT: Conceptualization, Methodology, Writing - Review \& editing, and Supervision.
CMH: Conceptualization, Methodology, Resources, Writing - Review \& editing, Supervision, and Funding acquisition.

\section*{AI Statement}
\noindent This paper has been proofread by a language model;  authors have read through the resulting content to ensure accuracy.

\clearpage
\balance

\bibliographystyle{plainnat}
\bibliography{references}

@inproceedings{huttenrauch2003help,
  title={To help or not to help a service robot},
  author={Huttenrauch, H and Eklundh, Kerstin Severinson},
  booktitle={The 12th IEEE International Workshop on Robot and Human Interactive Communication, 2003. Proceedings. ROMAN 2003.},
  pages={379--384},
  year={2003},
  organization={IEEE}
}

@article{innes2021experimental,
  title={Experimental studies of human--robot interaction: Threats to valid interpretation from methodological constraints associated with experimental manipulations},
  author={Innes, J Michael and W. Morrison, Ben},
  journal={International Journal of Social Robotics},
  volume={13},
  number={4},
  pages={765--773},
  year={2021}
}

@article{li2025hri,
  title={HRI-Confusion: A Multimodal Dataset for Modelling and Detecting User Confusion in Situated Human-Robot Interaction},
  author={Li, Na and Courtney, Jane and Ross, Robert},
  journal={Data in Brief},
  pages={112047},
  year={2025}
}

@article{edirisinghe2024security,
  title={Field Trial of a Queue-Managing Security Guard Robot},
  author={Edirisinghe, Sachi and Satake, Satoru and Liu, Yuyi and Kanda, Takayuki},
  journal={ACM Transactions on Human-Robot Interaction},
  volume={13},
  number={4},
  pages={1--48},
  year={2024}
}

@article{bu2025making,
  title={Making sense of robots in public spaces: A study of trash barrel robots},
  author={Bu, Fanjun and Fischer, Kerstin and Ju, Wendy},
  journal={ACM Transactions on Human-Robot Interaction},
  volume={14},
  number={4},
  pages={1--21},
  year={2025}
}

@inproceedings{brown2024trash,
author = {Brown, Barry and Bu, Fanjun and Mandel, Ilan and Ju, Wendy},
title = {Trash in Motion: Emergent Interactions with a Robotic Trashcan},
year = {2024},
booktitle = {Proceedings of the 2024 CHI Conference on Human Factors in Computing Systems},
numpages = {17},
}

@article{yamada2024qualitative,
  title={Qualitative Research of Robot-Helping Behaviors in a Field Trial},
  author={Yamada, Sachie and Kanda, Takayuki and Tomita, Kanako},
  journal={ACM Transactions on Human-Robot Interaction},
  volume={13},
  number={2},
  pages={1--19},
  year={2024}
}

@inproceedings{pineda2025see,
  title={“See You Later, Alligator”: Impacts of Robot Small Talk on Task, Rapport, and Interaction Dynamics in Human-Robot Collaboration},
  author={Pineda, Kaitlynn Taylor and Brown, Ethan and Huang, Chien-Ming},
  booktitle={2025 20th ACM/IEEE International Conference on Human-Robot Interaction (HRI)},
  pages={819--828},
  year={2025}
}

@inproceedings{michalowski2006spatial,
  title={A spatial model of engagement for a social robot},
  author={Michalowski, Marek P and Sabanovic, Selma and Simmons, Reid},
  booktitle={9th IEEE International Workshop on Advanced Motion Control, 2006.},
  pages={762--767},
  year={2006}
}

@article{sebo2020robots,
  title={Robots in groups and teams: a literature review},
  author={Sebo, Sarah and Stoll, Brett and Scassellati, Brian and Jung, Malte F},
  journal={Proceedings of the ACM on Human-Computer Interaction},
  volume={4},
  number={CSCW2},
  pages={1--36},
  year={2020}
}

@article{lee2024consumer,
  title={Consumer Attention to a Coffee Brewing Robot: An Eye-Tracking Study},
  author={Lee, Cho-Long and Kim, Sunmin and Lim, Manyoel and Hwang, Sungjae and Kim, Daekwang and Kwak, Han Sub},
  journal={Journal of Sensory Studies},
  volume={39},
  number={5},
  year={2024}
}

@inproceedings{kamino2023coffee,
  title={Coffee, tea, robots? the performative staging of service robots in'robot cafes' in japan},
  author={Kamino, Waki and Sabanovic, Selma},
  booktitle={Proceedings of the 2023 ACM/IEEE international conference on human-robot interaction},
  pages={183--191},
  year={2023}
}

@inproceedings{gucsi2025hrisense,
author = {Gucsi, Balint and Tan Viet Tuyen, Nguyen and Chu, Bing and Tarapore, Danesh and Tran-Thanh, Long},
title = {HRI-SENSE: A Multimodal Dataset on Social and Emotional Responses to Robot Behaviour},
year = {2025},
booktitle = {Proceedings of the 2025 ACM/IEEE International Conference on Human-Robot Interaction},
pages = {1319–1323}
}

@inproceedings{chan2025field,
author = {Chan, Yao-Cheng},
title = {A Field Observation of Incidental Human-Robot Encounters in Public},
year = {2025},
booktitle = {Proceedings of the 2025 ACM/IEEE International Conference on Human-Robot Interaction},
pages = {1265–1268}
}

@inproceedings{schiffmann2025predicting,
author = {Schiffmann, Michael and Chojnowski, Oliver and Richert, Anja},
title = {Predicting User Satisfaction in a Public Space HRI-Scenario},
year = {2025},
booktitle = {Proceedings of the 2025 ACM/IEEE International Conference on Human-Robot Interaction},
pages = {1598–1602}
}

@inproceedings{tabatabaei2025real,
  title={Real-Time Detection of Robot Failures Using Gaze Dynamics in Collaborative Tasks},
  author={Tabatabaei, Ramtin and Kostakos, Vassilis and Johal, Wafa},
  booktitle={2025 20th ACM/IEEE International Conference on Human-Robot Interaction (HRI)},
  pages={1660--1664},
  year={2025}
}

@inproceedings{tabatabaei2025gazing,
  title={Gazing at failure: Investigating human gaze in response to robot failure in collaborative tasks},
  author={Tabatabaei, Ramtin and Kostakos, Vassilis and Johal, Wafa},
  booktitle={2025 20th ACM/IEEE International Conference on Human-Robot Interaction (HRI)},
  pages={939--948},
  year={2025}
}

@inproceedings{parreira2024bad,
  title={“Bad Idea, Right?” Exploring Anticipatory Human Reactions for Outcome Prediction in HRI},
  author={Parreira, Maria Teresa and Lingaraju, Sukruth Gowdru and Ramirez-Artistizabal, Adolfo and Bremers, Alexandra and Saha, Manaswi and Kuniavsky, Michael and Ju, Wendy},
  booktitle={2024 33rd IEEE International Conference on Robot and Human Interactive Communication (ROMAN)},
  pages={2072--2078},
  year={2024}
}

@inproceedings{cao2025err,
    author = {Cao, Shiye and Stiber, Maia and Mahmood, Amama and Parreira, Maria Teresa and Ju, Wendy and Spitale, Micol and Gunes, Hatice and Huang, Chien-Ming},
    title = {ERR@ HRI 2.0 Challenge: Multimodal Detection of Errors and Failures in Human-Robot Conversations},
    booktitle = {Proceedings of the 33rd ACM International Conference on Multimedia (ACM MM)} ,
    year = {2025}}

@inproceedings{spitale2024err,
  title={Err@ hri 2024 challenge: Multimodal detection of errors and failures in human-robot interactions},
  author={Spitale, Micol and Parreira, Maria Teresa and Stiber, Maia and Axelsson, Minja and Kara, Neval and Kankariya, Garima and Huang, Chien-Ming and Jung, Malte and Ju, Wendy and Gunes, Hatice},
  booktitle={Proceedings of the 26th International Conference on Multimodal Interaction},
  pages={652--656},
  year={2024}
}

@INPROCEEDINGS{spitale2023longitudinal,
  author={Spitale, Micol and Axelsson, Minja and Kara, Neval and Gunes, Hatice},
  booktitle={2023 32nd IEEE International Conference on Robot and Human Interactive Communication (RO-MAN)}, 
  title={Longitudinal Evolution of Coachees’ Behavioural Responses to Interaction Ruptures in Robotic Positive Psychology Coaching}, 
  year={2023},
  pages={315-322}}

@inproceedings{spitale2023robotic,
author = {Spitale, Micol and Axelsson, Minja and Gunes, Hatice},
title = {Robotic Mental Well-being Coaches for the Workplace: An In-the-Wild Study on Form},
year = {2023},
booktitle = {Proceedings of the 2023 ACM/IEEE International Conference on Human-Robot Interaction},
pages = {301–310}
}

@INPROCEEDINGS{liu2025done,
  author={Liu, Shannon and Parreira, Maria Teresa and Ju, Wendy},
  booktitle={2025 20th ACM/IEEE International Conference on Human-Robot Interaction (HRI)}, 
  title={“I'm Done”: Describing Human Reactions to Successive Robot Failure}, 
  year={2025},
  volume={},
  number={},
  pages={1458-1462}
}

@article{andrist2017what,
author = {Andrist, Sean and Bohus, Dan and Kamar, Ece and Horvitz, Eric},
title = {What Went Wrong and Why? Diagnosing Situated Interaction Failures in the Wild},
year = {2017},
booktitle = {ICSR 2017: Social Robotics},
pages={293-303}
}

@article{jung2018robots,
author = {Jung, Malte and Hinds, Pamela},
title = {Robots in the Wild: A Time for More Robust Theories of Human-Robot Interaction},
year = {2018},
issue_date = {May 2018},
publisher = {Association for Computing Machinery},
address = {New York, NY, USA},
volume = {7},
number = {1},
doi = {10.1145/3208975},
journal = {J. Hum.-Robot Interact.}
}

@INPROCEEDINGS{sabanovic2006robots,
  author={Sabanovic, S. and Michalowski, M.P. and Simmons, R.},
  booktitle={9th IEEE International Workshop on Advanced Motion Control, 2006.}, 
  title={Robots in the wild: observing human-robot social interaction outside the lab}, 
  year={2006},
  pages={596-601},
  doi={10.1109/AMC.2006.1631758}}

@inproceedings{koike2025what,
      title={What Drives You to Interact?: The Role of User Motivation for a Robot in the Wild}, 
      author={Amy Koike and Yuki Okafuji and Kenya Hoshimure and Jun Baba},
      booktitle={Proceedings of the 2025 ACM/IEEE International Conference on Human-Robot Interaction},
      year={2025}
}

@article{edirisinghe2024field,
  title={Field Trial of a Queue-Managing Security Guard Robot},
  author={Edirisinghe, Sachi and Satake, Satoru and Liu, Yuyi and Kanda, Takayuki},
  journal={ACM Transactions on Human-Robot Interaction},
  volume={13},
  number={4},
  pages={1--48},
  year={2024}
}

@inproceedings{belcamino2024gaze,
  title={Gaze-Based Intention Recognition for Human-Robot Collaboration},
  author={Belcamino, Valerio and Takase, Miwa and Kilina, Mariya and Carf{\`\i}, Alessandro and Mastrogiovanni, Fulvio and Shimada, Akira and Shimizu, Sota},
  booktitle={Proceedings of the 2024 International Conference on Advanced Visual Interfaces},
  pages={1--5},
  year={2024}
}

@inproceedings{sauppe2014social,
  title={How social cues shape task coordination and communication},
  author={Saupp{\'e}, Allison and Mutlu, Bilge},
  booktitle={Proceedings of the 17th ACM conference on Computer supported cooperative work \& social computing},
  pages={97--108},
  year={2014}
}

@inproceedings{vinciarelli2008social,
  title={Social signals, their function, and automatic analysis: a survey},
  author={Vinciarelli, Alessandro and Pantic, Maja and Bourlard, Herv{\'e} and Pentland, Alex},
  booktitle={Proceedings of the 10th international conference on Multimodal interfaces},
  pages={61--68},
  year={2008}
}

@article{candon2023nonverbal,
  title={Nonverbal human signals can help autonomous agents infer human preferences for their behavior},
  author={Candon, Kate and Chen, Jesse and Kim, Yoony and Tsoi, Nathan and V{\'a}zquez, Marynel},
  year={2023},
  publisher={ACM}
}

@article{duric2002integrating,
  title={Integrating perceptual and cognitive modeling for adaptive and intelligent human-computer interaction},
  author={Duric, Zoran and Gray, Wayne D and Heishman, Ric and Li, Fayin and Rosenfeld, Azriel and Schoelles, Michael J and Schunn, Christian and Wechsler, Harry},
  journal={Proceedings of the IEEE},
  volume={90},
  number={7},
  pages={1272--1289},
  year={2002},
  publisher={IEEE}
}

@article{stiber2025robot,
  title={Robot Error Awareness Through Human Reactions: Implementation, Evaluation, and Recommendations},
  author={Stiber, Maia and Taylor, Russell and Huang, Chien-Ming},
journal={arXiv preprint arXiv:2501.05723},
  year={2025}
}

@inproceedings{ravishankar2024zero,
  title={Zero-Shot Learning to Enable Error Awareness in Data-Driven HRI},
  author={Ravishankar, Joshua and Doering, Malcolm and Kanda, Takayuki},
  booktitle={Proceedings of the 2024 ACM/IEEE International Conference on Human-Robot Interaction},
  pages={592--601},
  year={2024}
}

@inproceedings{severitt2024communication,
  title={Communication breakdown: Gaze-based prediction of system error for AI-assisted robotic arm simulated in VR},
  author={Severitt, Bj{\"o}rn Rene and Lenhart, Patrizia and Hosp, Benedikt Werner and Castner, Nora Jane and Wahl, Siegfried},
  booktitle={Proceedings of the 2024 Symposium on Eye Tracking Research and Applications},
  pages={1--7},
  year={2024}
}

@inproceedings{stiber2024uh,
  title={``Uh, \emph{This} One?'': Leveraging Behavioral Signals for Detecting Confusion during Physical Tasks},
  author={Stiber, Maia and Bohus, Dan and Andrist, Sean},
  booktitle={International Conference on Multimodal Interaction},
  year={2024}
}

@inproceedings{bremers2023bystander,
  title={The Bystander Affect Detection (BAD) Dataset for Failure Detection in HRI},
  author={Bremers, Alexandra and Parreira, Maria Teresa and Fang, Xuanyu and Friedman, Natalie and Ramirez-Aristizabal, Adolfo and Pabst, Alexandria and Spasojevic, Mirjana and Kuniavsky, Michael and Ju, Wendy},
  booktitle={2023 IEEE/RSJ International Conference on Intelligent Robots and Systems (IROS)},
  pages={11443--11450},
  year={2023},
  organization={IEEE}
}

@inproceedings{lewis2009using,
  title={Using humans as sensors in robotic search},
  author={Lewis, Michael and Wang, Huadong and Velagapudi, Prasanna and Scerri, Paul and Sycara, Katia},
  booktitle={2009 12th International Conference on Information Fusion},
  pages={1249--1256},
  year={2009},
  organization={IEEE}
}

@inproceedings{wolf2021gaze,
  title={Gaze comes in handy: Predicting and preventing erroneous hand actions in ar-supported manual tasks},
  author={Wolf, Julian and Lohmeyer, Quentin and Holz, Christian and Meboldt, Mirko},
  booktitle={2021 IEEE International Symposium on Mixed and Augmented Reality (ISMAR)},
  pages={166--175},
  year={2021},
  organization={IEEE}
}

@inproceedings{scherf2024you,
  title={Are You Sure?-Multi-Modal Human Decision Uncertainty Detection in Human-Robot Interaction},
  author={Scherf, Lisa and Gasche, Lisa Alina and Chemangui, Eya and Koert, Dorothea},
  booktitle={Proceedings of the 2024 ACM/IEEE International Conference on Human-Robot Interaction},
  pages={621--629},
  year={2024}
}

@article{cuadra2021my,
  title={My bad! repairing intelligent voice assistant errors improves interaction},
  author={Cuadra, Andrea and Li, Shuran and Lee, Hansol and Cho, Jason and Ju, Wendy},
  journal={Proceedings of the ACM on Human-Computer Interaction},
  volume={5},
  number={CSCW1},
  pages={1--24},
  year={2021},
  publisher={ACM New York, NY, USA}
}

@article{bremers2023using,
  title={Using Social Cues to Recognize Task Failures for HRI: A Review of Current Research and Future Directions},
  author={Bremers, Alexandra and Pabst, Alexandria and Parreira, Maria Teresa and Ju, Wendy},
  journal={arXiv preprint arXiv:2301.11972},
  year={2023}
}

@inproceedings{stiber2023using,
  title={On using social signals to enable flexible error-aware HRI},
  author={Stiber, Maia and Taylor, Russell H and Huang, Chien-Ming},
  booktitle={Proceedings of the 2023 ACM/IEEE International Conference on Human-Robot Interaction},
  pages={222--230},
  year={2023}
}

@article{ben2019early,
  title={Early detection of user engagement breakdown in spontaneous human-humanoid interaction},
  author={Ben-Youssef, Atef and Clavel, Chlo{\'e} and Essid, Slim},
  journal={IEEE Transactions on Affective Computing},
  year={2019}
}

@article{tabrez2020survey,
  title={A Survey of Mental Modeling Techniques in Human--Robot Teaming},
  author={Tabrez, Aaquib and Luebbers, Matthew B and Hayes, Bradley},
  journal={Current Robotics Reports},
  year={2020}
}

@article{stiber2022Forging,
    title = {Forging Productive Human-Robot Partnerships Through Task Training},
	Author = {Stiber, Maia and Gao, Yuxiang and Taylor, Russell H. and Huang, Chien-Ming},
	journal = {ACM Transactions on Human Robot-Interaction},
	year = {2022}}

@inproceedings{stiber2022modeling,
	Author = {Stiber, Maia and Taylor, Russell H. and Huang, Chien-Ming},
	Booktitle = {IEEE/RSJ International Conference on Intelligent Robots and Systems},
	Title = {Modeling Human Response to Robot Errors for Timely Error Detection},
	Year = {2022}}

@article{hassin2013inherently,
  title={Inherently ambiguous: Facial expressions of emotions, in context},
  author={Hassin, Ran R and Aviezer, Hillel and Bentin, Shlomo},
  journal={Emotion Review},
  volume={5},
  number={1},
  year={2013}
}

@misc{bohus2021platform,
      title={Platform for Situated Intelligence}, 
      author={Dan Bohus and Sean Andrist and Ashley Feniello and Nick Saw and Mihai Jalobeanu and Patrick Sweeney and Anne Loomis Thompson and Eric Horvitz},
      year={2021},
      eprint={2103.15975},
      archivePrefix={arXiv},
      primaryClass={cs.AI}
}

@inproceedings{stiber2020not, 
    author = {Stiber, Maia and Huang, Chien-Ming}, title = {Not All Errors Are Created Equal: Exploring Human Responses to Robot Errors with Varying Severity}, year = {2020}, booktitle = {Companion Publication of the 2020 International Conference on Multimodal Interaction}, pages = {97–101}}

@inproceedings{kontogiorgos2021systematic,
	author = {Kontogiorgos, Dimosthenis and Tran, Minh and Gustafson, Joakim and Soleymani, Mohammad},
	booktitle = {Proceedings of the 2021 International Conference on Multimodal Interaction},
	title = {A Systematic Cross-Corpus Analysis of Human Reactions to Robot Conversational Failures},
	year = {2021}}

@inproceedings{trung2017head,
	Author = {Pauline Trung and Manuel Giuliani and Markus Miksch and Gerald Stollnberger and Susanne Stadler and Nicole Mirnig and Manfred Tscheligi},
	Booktitle = {ACM International Conference on Multimodal Interaction},
	Title = {Head and Shoulders: Automatic Error Detection in Human-Robot Interaction},
	Year = {2017}}

@inproceedings{aronson2018gaze,
	Author = {Reuben M. Aronson and Henny Admoni},
	Booktitle = {RSS Workshop: Towards a Framework for Joint Action},
	Title = {Gaze for Error Detection During Human-Robot Shared Manipulation},
	Year = {2018}}

@article{mirnig2017err,
	Author = {Nicole Mirnig and Gerald Stollnberger and Markus Miksch and Susanne Stadler and Manuel Giuliani and Manfred Tscheligi},
	Journal = {Frontiers in Robotics and AI},
	Title = {To Err Is Robot: How Humans Assess and Act toward an Erroneous Social Robot},
	Volume = {4},
	Year = {2017}}

@inproceedings{cahya2019static,
	Author = {Dito Eka Cahya and Rahul Ramakrishnan and Manuel Giuliani},
	Booktitle = {International Conference on Social Robotics},
	Pages = {189--199},
	Title = {Static and Temporal Differences in Social Signals Between Error-Free and Erroneous Situations in Human-Robot Collaboration},
	Year = {2019}}

@article{giuliani2015systematic,
	Author = {Manuel Giuliani and Nicole Mirnig and Gerald Stollnberger and Susanne Stadler and Roland Buchner and Manfred Tscheligi},
	Journal = {Frontiers in Psychology},
	Title = {Systematic Analysis of Video Data from Different Human-Robot Interaction Studies: A Categorisation of Social Signals During Error Situations},
	Volume = {6},
	Year = {2015}}

@inproceedings{mirnig2015impact,
	Author = {Nicole Mirnig and Manuel Giuliani and Gerald Stollnberger and Susanne Stadler and Roland Buchner and Manfred Tscheligi},
	Booktitle = {International Conference on Social Robotics},
	Organization = {Springer},
	Pages = {461--471},
	Title = {Impact of Robot Actions on Social Signals and Reaction Times in HRI Error Situations},
	Year = {2015}}

\clearpage

\section*{Appendix 1: Field Deployment and Data Collection System Code}
The coffee robot field deployment and data collection system (described in Section 3) used in the study is open-source and be found at this link: [Will be released when published].

\paragraph*{Additional Component---LLM-Powered Interaction}
An additional component this system has that was not used in the deployment is the \emph{LLM-Powered Interaction} component. It allows the user to communicate to the robot verbally and/or the robot to reason about the interaction and its behavior so the robot can respond verbally and behaviorally. It can send and receive commands from the \emph{Ordering Interface} and \emph{Robot Controller} and interfaces with the speaker. It can be toggled on or off. For the sake of our data collection field deployment, we did not use this component. There are two versions of this component: single LLM and LLM chain.

\textbf{Single LLM.}
The single LLM is set up for verbal inputs from the system's microphone, converted to text and then fed in. The single LLM also takes in the system prompt and then outputs text that is parsed and fed to a speech generator, enabling the robot to speak.

\textbf{LLM Chain.}
The LLM chain allows for the chaining of multiple LLMs with defined transition states (transition states can be modified). It takes input from user speech as well as robot status to inform initiation and transitions among the defined LLMs. The LLMs can run sequentially and loop (both within one LLM and back to a previous one) as determined by predefined logic or by the LLMs (if prompted). The output is text that can be used to trigger speech, robot behavior, and other defined behavior as needed.

\section*{Appendix 2: Codebook}
Below is the codebook (Table~\ref{tab:codebook}) used to annotate user behavioral patterns and social signals expressed when interacting with the coffee robot during field deployment and in the face of robot errors.

\newpage
\nobalance

\begin{landscape}
\begin{table}[]
\centering
\caption{Codebook for coffee robot field deployment.}
\label{tab:codebook}
\begin{tabular}{p{4cm}rp{6.5cm}p{5cm}}
\textbf{High-Level Code Categories} & \textbf{Theme} & \textbf{Codes} & \textbf{Examples} \\ \hline \hline
Robot Error Codes & Error Presence (binary) & Yes/No &  \vspace{0.3cm}\\
 & Error Temporal & Error Start & \vspace{0.3cm} \\
 & Reacted to? (binary) & Yes/No & \vspace{0.3cm} \\
 & Error Expectedness (binary) & Preplanned/Unexpected & \vspace{0.2cm} \\ \hline
Behavioral Codes & Social Signal Responses (presence) & Smiling; Laughing; Talking to robot; Talking to group member; Talking to self; Talking to passersby; Non-lexical Utterances; Using phone; Recording Robot; Looking away; Walking around the deployment; Walking away; Leaning in; Grimacing; Shocked expression; Shaking head; Freezing; Gesturing with hands; Physically assisting the robot &  \vspace{2.6cm}\\
 & Reaction to Error Temporal & Reaction Start & \\
 & & Reaction Stop & \vspace{0.3cm} \\
 & Reaction Stimuli & Robot; Robot Error; Robot Error Propagation; Other group members; Passerby; Phone & \vspace{0.7cm} \\ 
 & Robot Relevant Verbal Behavior Valence & Positive/Neutral/Negative & \\ \hline
Verbal Level of Relevancy and & Error-related (sentence-level) & 0: not related to robot error; & \eg. \textit{``Thank you''} \\
Information Codes &  & 1: instinctual reaction to error/indicator of error presence; & \eg \textit{``Oh, no no no''} \\
 &  & 2: describes the symptom (end result of the error)--\emph{volunteered information}; & \eg \textit{``Hey, again, it missed the coffee.''} \\
 &  & 3: hypothesizes about cause or active debugging---\emph{volunteered information}; & \eg \textit{``Is this the problem of this coffee machine?''} \vspace{0.7cm}\\
 & Robot/Task-related (sentence-level) & 0: not related to robot or task; & \eg \textit{``Happy Birthday''} \\
 &  & 1: \emph{describes} current robot behavior or task state (facts) & \eg \textit{``Oh, it turned it off.''} \\
 &  & 2: \emph{evaluates} current robot behavior (providing feedback and error information) or \emph{reflects} on past interaction---\emph{volunteered information} & \eg \textit{``Like, this is the hardest part.''} \\
 &  & 3: \emph{speculates} on robot behavior/ capabilities/ intent/ process---\emph{volunteered information} & \eg \textit{``I thought I was just gonna hand it to you all erratically.''} \\ \hline
\end{tabular}%
\end{table}
\end{landscape}

\end{document}